\documentclass[preprints,article,accept,moreauthors,pdftex,10pt,a4paper]{mdpi}

\usepackage{graphicx}
\usepackage{wrapfig}
\usepackage{caption}
\usepackage{subcaption}
\usepackage{multirow}
\usepackage{amsmath}
\usepackage{mathtools}
\usepackage{amsfonts}
\usepackage[export]{adjustbox}
\usepackage{stfloats}
\usepackage{tabularx}

\firstpage{1} 
\pubvolume{xx}
\issuenum{1}
\articlenumber{5}
\pubyear{2018}
\copyrightyear{2018}
\history{Received: date; Accepted: date; Published: date}

\pdfoutput=1

\Title{Real-time Monocular Visual Odometry for Turbid and Dynamic Underwater Environments}

\Author{Maxime Ferrera $^{1,2}$, Julien Moras $^{1}$, Pauline Trouv\'e-Peloux$^{1}$ and Vincent Creuze$^{2}$}

\AuthorNames{Maxime Ferrera, Julien Moras, Pauline Trouv\'e-Peloux and Vincent Creuze}

\address{%
$^{1}$ \quad DTIS, ONERA, Universit\'e Paris Saclay F-91123 Palaiseau - France; maxime.ferrera@onera.fr (M.F.); julien.moras@onera.fr (J.M.); pauline.trouve@onera.fr (P.T.) \\
$^{2}$ \quad LIRMM, Univ. Montpellier, CNRS, Montpellier, France; vincent.creuze@lirmm.fr (V.C.)}

\corres{Correspondence: maxime.ferrera@onera.fr}

\abstract{In the context of underwater robotics, the visual degradation induced by the medium properties make difficult the exclusive use of cameras for localization purpose.  Hence, many underwater localization methods are based on expensive navigation sensors associated with acoustic positioning.  On the other hand, pure visual localization methods have shown great potential in underwater localization but the challenging conditions, such as the presence of turbidity and dynamism, remain complex to tackle.\\
In this paper we propose a new visual odometry method designed to be robust to these visual perturbations.  The proposed algorithm has been assessed on both simulated and real underwater datasets and outperforms state-of-the-art terrestrial visual SLAM methods under many of the most challenging conditions.  The main application of this work is the localization of Remotely Operated Vehicles used for underwater archaeological missions, but the developed system can be used in any other applications as long as visual information is available.}


\begin{document}

\section{Introduction}
\label{sec:intro}

Accurate localization is critical for most robotic underwater operations, especially when navigating in areas with obstacles such as rocks, shipwrecks or Oil \& Gas structures.  In underwater archaeology, \textit{Remotely Operated Vehicles} (ROVs) are used to explore and survey sites in deep waters.  Even if these robots are remotely operated by a pilot, systems providing their real-time localization are valuable to efficiently use them.  For example, this information can be used to ensure the completeness of photogrammetric surveys or as a feedback for autonomous navigation.  

As radio signals are absorbed by sea water, it is not possible to use GPS systems to localize underwater vehicles.  Acoustic positioning systems like \textit{Ultra Short Baseline} (USBL), \textit{Short Baseline} or \textit{Long Baseline} (LBL) can be used as GPS alternatives.  However these systems are expensive and require precise calibration to get a positioning accuracy in the order of one meter.  In order to obtain submetric accuracy, the use of high-end \textit{Inertial Navigation Systems} (INS) and of \textit{Doppler Velocity Logs} (DVL) is often necessary.  Some of the existing approaches for underwater localization complement these sensors with sonars to limit the drift due to the integration of measurements errors \cite{Paull2014AUVReview}.  Such setups are mandatory if one seeks to localize underwater vehicles in the middle of the water column.  However, when navigating close to the seabed, visual information becomes available given that the vehicle embeds a lightning system.  In this scenario, cameras have also been used as a complementary sensor to limit the drift by matching temporally spaced images \cite{EusticeVAN,JohnsonRoberson3Dreconstruction,MahonEfficientViewBased,Beall_isam,Bi_objective_BA,stereo_graph_slam}.  If the aforementioned approaches have shown good results on very large trajectories, they require the use of expensive high-end navigational sensors as the cameras or the acoustic positioning systems are only used to constrain the drift.
robots motion as the cameras or the acoustic positioning systems are only used to constrain the drift.

\begin{figure}[!t]
	\centering{
	\includegraphics[width=0.6\linewidth]{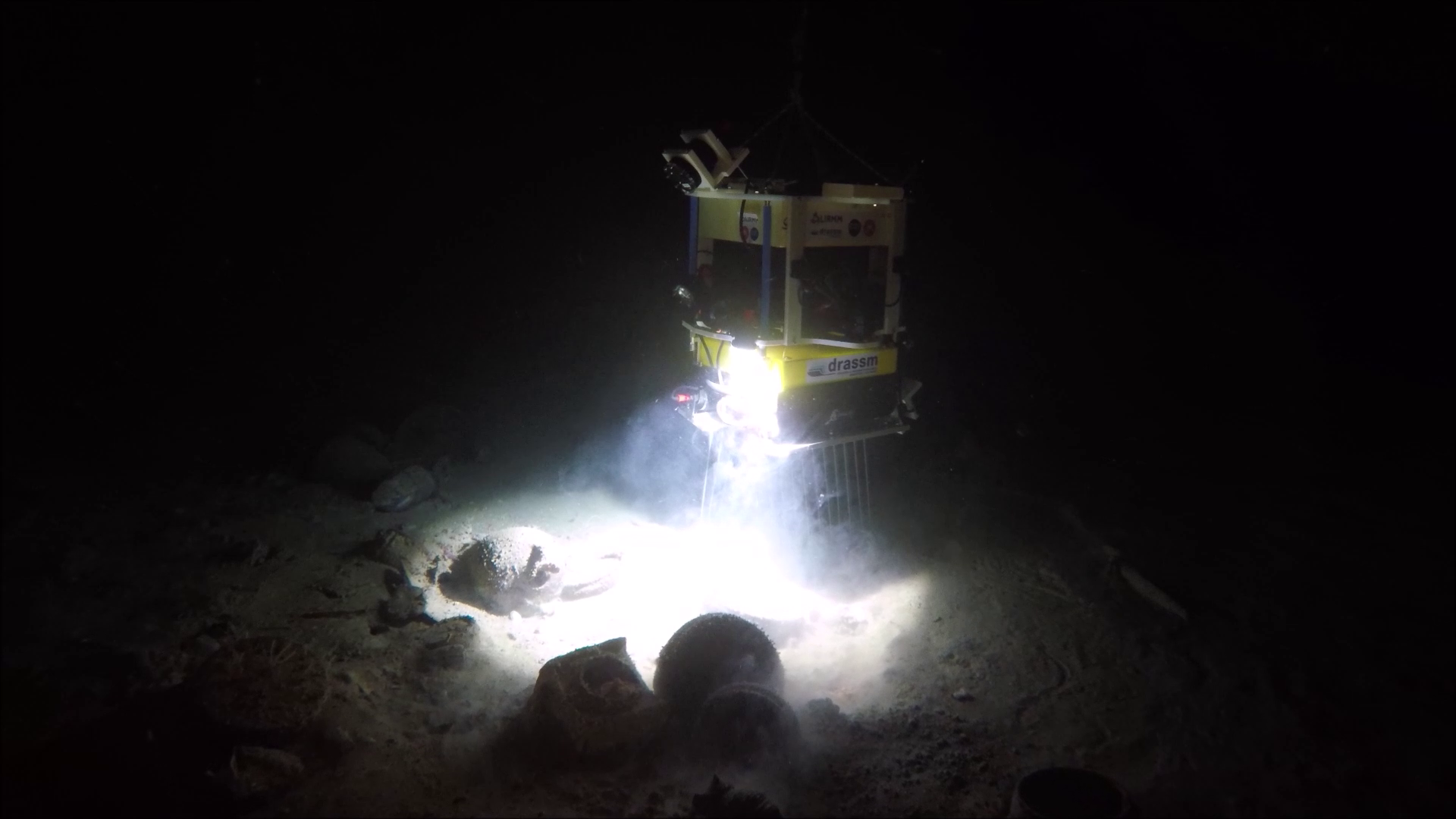}
    }
    \setlength{\belowcaptionskip}{-15pt}
\caption{\small Remotely Operated Vehicle \textit{Dumbo} performing archaeological sampling on an antic shipwreck in the Mediterranean Sea (460 meters deep). This archaeological operation has been performed under the supervision of the French Department of Underwater Archaeology (DRASSM), in accordance with the UNESCO Convention on the Protection of the Underwater Cultural Heritage. Credit: F. Osada, DRASSM - Images Explorations.}
    \label{fig:intro}
\end{figure}

In contrast, in this work we are interested in the development of a submeter grade localization system from a minimal set of low-cost sensors for lightweight ROVs used for deep archaeological operations (Fig.\ref{fig:intro}).  As an ROV always embeds a camera for remote control purpose, we decided to develop a visual localization framework based solely on a monocular camera to estimate in real-time the ego-motion of the robot.
  
Visual Odometry (VO) and Visual Simultaneous Localization And Mapping (VSLAM) have been a great topic of research over the past decades \cite{PastPresentFutureSLAM}.  VSLAM differs from VO by maintaining a reusable global map, allowing the detection of loop closures when seeing again already mapped scenes.  In underwater environment, localization from vision is more complex than in aerial and terrestrial environments and state-of-the-art open-source VO or VSLAM algorithms fail when the operating conditions become too harsh \cite{ClusterBased_UW_LC,UW_Cave_Mapping}.
This is mainly due to the visual degradation caused by the medium specific properties.  Indeed, the strong light absorption of the medium shortens the visual perception to a few meters and makes the presence of an artificial lightning system mandatory when operating in deep waters.  Besides, the propagation of light is backscattered by floating particles, causing turbidity effects on the captured images.  In the darkness of deep waters, the fauna is also a cause of visual degradation as animals are attracted by the artificial light and tend to get in the field of view of the camera, leading to dynamism and occlusions in the images.  In front of these difficulties, many works tackle the underwater localization problem using sonar systems \cite{MarinaSLAM,malta_sonar-slam,aekf-slam}, as they do not suffer from these visual degradation.  Nevertheless, the information delivered by a sonar is not as rich as optical images \cite{BONINFONT2015-OceanEngineering} and remains very challenging to analyze.  Furthermore, at close range, acoustic systems do not provide accurate enough localization information whereas visual sensing can be highly effective \cite{PALOMERAS2018-AUV-Docking}.

In this paper we propose UW-VO (UnderWater-Visual Odometry), a new monocular VO algorithm dedicated to the underwater environment that overcomes the aforementioned visual degradation.  Inspired by state-of-the-art VO and VSLAM methods from aerial and terrestrial robotics, UW-VO is a keyframe based method.  The front-end of UW-VO has been made robust to turbidity and short occlusions by employing optical flow to track features and a retracking mechanism to retrieve temporarily lost features.  The back-end relies on Bundle Adjustment \cite{BundleAdjustment} to ensure minimal drift and consistency in the reconstructed trajectory.

\pagebreak

\noindent The paper contributions are the following:

\begin{itemize}
\item A thorough evaluation of visual features tracking methods on underwater images.
\item The development of UW-VO: a monocular VO method robust to turbidity and to short occlusions caused by the environment dynamism (animals, algae...).
\item An evaluation of state-of-the-art open-source monocular VO and VSLAM algorithms on underwater datasets and a comparison to UW-VO, highlighting its robustness to underwater visual degradation.
\item The release of a dataset consisting of video sequences taken on an underwater archaeological site\footnote{https://seafile.lirmm.fr/d/aa84057dc29a4af8ae4a/}. 
\end{itemize}

The paper is then organized as follows.  First, in section \ref{sec:sota}, we review works related to our contributions.  In section \ref{sec:track_eval}, we present an evaluation of features tracking methods on two underwater sets of images and we show that optical flow based tracking performs better than methods based on the matching of descriptors.  In section \ref{sec:vo_framework}, our monocular visual localization system UW-VO is described.  Finally, in section \ref{sec:results}, we compare UW-VO to state-of-the-art open-source monocular VO and VSLAM on both a synthetic dataset and on a real one that we made publicly available for the community.  We show that UW-VO competes with these methods in terms of accuracy and surpasses them in terms of robustness.

\section{Related Work}
\label{sec:sota}

The localization of robots from the output of a camera system has been a great topic of research for several decades.  Here we review the major works in the field of underwater visual localization and we compare it to the aerial and terrestrial robotics field. \\

\noindent \textbf{Underwater Features Tracking.}  The problem of features tracking is one of the cornerstones of visual localization.  In the underwater context, \cite{FeatExtract4UwSLAM} evaluate the use of SURF \cite{Surf} features in their SLAM algorithms and assess improvements in the localization but their method do not run online.  The authors of \citet{Shkurti_FeatEval4PoseEstim} evaluate different combinations of feature detectors and descriptors and they show that SURF features and Shi-Tomasi corners \cite{GFTT} matched by \textit{Zero Mean Normalized Cross-Correlation} (ZNCC) are good combinations for visual state estimation.  None of these works included the evaluation of binary features such as BRIEF \cite{Brief} or ORB \cite{Orb} nor of optical-flow, which are widely used methods for features tracking in VO and VSLAM.
In \cite{UwFeatDetection_Garcia,Turbid_Dataset_2015}, many feature detectors are evaluated on underwater images with a focus on their robustness to turbidity.  They follow the evaluation protocol of \cite{Mikolajczyk2005} by using the detectors repeatability as their metric.  Robustness to turbidity is essential for underwater visual localization but only evaluating repeatability of the detectors does not ensure good features tracking capacity as ambiguity can still arise when trying to match these features between two images.  Our features tracking evaluation differs from these previous works by directly evaluating the features tracking efficiency of a wide range of feature detectors and tracking methods in the context of VO and VSLAM. \\

\noindent \textbf{Underwater Visual Localization.}  \citet{EusticeVAN} were among the first to present a successful use of visual information as a complementary sensor for underwater robots localization.  They used an Extended Information Filter (EIF) to process dead-reckoning sensors and insert visual constraints based on the 2D-2D matching of points coming from overlapping monocular images.  Here, only the relative pose between cameras is computed so the visual motion is estimated up to scale and do not provide the full 6 degrees of freedom of the robot motions.   
Following their work, many stereo-vision based systems were proposed \cite{MahonEfficientViewBased,JohnsonRoberson3Dreconstruction,Pfingsthorn16}.  The advantage of using stereo cameras lies in the availability of the metric scale in opposition to the scale ambiguity inherent to pure monocular system.  The scale factor can indeed be resolved from the known baseline between both sensors (assuming the stereo system calibrated).  
\cite{Beall_isam,Bi_objective_BA,stereo_graph_slam} later integrated nonlinear optimization steps to further process the visual data through bundle adjustment \cite{BundleAdjustment}.  \cite{VisualSaliencySLAM} extended \cite{EusticeVAN} by keeping a monocular approach but adding loop-closure capability to their methods through the computation of a visual saliency metric using SIFT features.  However, in all these methods the visual information is only used to bound the localization drift using low-overlap imagery systems (1-2hz), but their systems mainly rely on expensive navigational sensors.  

Closer to our work, some stereo VO approaches use higher frame rate videos (10-20 hz) to estimate underwater vehicles ego-motion \cite{Corke07,DrapPhotogrammetry,SelectiveUWstereoVO}.  In \cite{DrapPhotogrammetry}, features are matched within stereo pairs to compute 3D point clouds and the camera poses are estimated by aligning these successive point clouds, making it a pure stereo vision method.  In parallel, the authors of \cite{SelectiveUWstereoVO} use a keyframe-based approach but their features tracking is done by matching descriptors both spatially (between stereo images pair) and temporally.  Moreover, they do not perform bundle adjustment to optimize the estimated trajectory.   

Despite the advantage of stereo-vision systems over monocular cameras, embedding a single camera is materially more practical, as classical camera housings can be used and cameras synchronization issues are avoided.  Furthermore, developing a monocular VO algorithm makes it portable to any kind of underwater vehicles, as long as it is equipped with a camera.  Even if there is a projective ambiguity with monocular systems,  it is possible to retrieve the scale factor from any complementary sensor capable of measuring a metric quantity. 

The early works of \cite{Garcia,Gracias} studied the use of a monocular camera as a mean of motion estimations for underwater vehicles navigating near the seabed.  In \cite{Garcia}, low-overlap monocular images are used to estimate the robot motions but the processing is performed offline.  \citet{Gracias} proposed a real-time mosaic-based visual localization method, estimating the robot motions through the computation of homographies with the limiting assumptions of purely planar scenes and 4 degrees of freedom motions ($x, y, z ,yaw$).  \cite{Negahdaripour06} extended these by computing the 6 degrees of freedom of a camera equipped with inclinometers.  In \cite{Nicosevici09}, an offline Structure From Motion framework was proposed to compute high quality 3D mosaicing of underwater environment from monocular images.

Underwater monocular-based methods using cameras at high frame rate (10-20hz) were studied by \cite{ShkurtiEKF-SLAM} and \cite{TrajBasedVisualSLAM}.  In their approaches, they fuse visual motion estimation in an Extended Kalman Filter (EKF) along with an IMU and a pressure sensor.  By using an EKF, they suffer from the integration of linearization errors which are limited in our system thanks to the iterative structure of bundle adjustment.  The authors of \cite{Carreras} make use of a camera to detect known patterns in a structured underwater environment and use it to improve the localization estimated by navigation sensors integrated into an EKF.  However, such methods are limited to known and controlled environment.  More recently, \cite{CreuzeMonoVO} presented a monocular underwater localization method that does not rely on an EKF framework but iteratively estimates ego-motion by integration of optical flow measurements corrected by an IMU and a pressure sensor. This latter is used to compute the scale factor of the observed scene. \\

\noindent \textbf{Aerial and Terrestrial Visual Localization.}  While most of the underwater odometry or SLAM systems rely on the use of an EKF, or its alternative EIF version, in aerial-terrestrial SLAM, filtering methods have been put aside to the profit of more accurate keyframe-based approaches using bundle adjustment \cite{WhyFilter}.  PTAM \cite{PTAM} was one of the first approach able to use bundle adjustment in real-time along with \cite{mouragnon}.  The work of \citet{Strasdat_dwo} and  ORB-SLAM from \citet{ORB-SLAM} are built on PTAM and improve it by adding a loop-closure feature highly reducing the localization drift by detecting loops in the trajectories.  Whereas all these methods match extracted features between successive frames, SVO \cite{SVO-2} and LSD-SLAM \cite{LSD-SLAM} are two direct approaches directly tracking photometric image patches to estimate the camera motions.  Following these pure visual systems, tightly coupled visual-inertial systems have been recently presented \cite{Leutenegger-OKVIS_ijrr,Rovio,Vins-Mono} with higher accuracy and robustness than standard visual systems.  These visual-inertial systems are all built on very accurate pure visual SLAM or VO methods, as they use low-cost \textit{Micro Electro Mechanical Systems (MEMS)} IMU, highly prone to drift.
 
Before considering the coupling of such complementary low-cost sensors for localization, the first step is to be able to rely on an accurate VO method.  Hence, contrarily to most of the approaches in underwater localization, we propose here a keyframe-based VO method, solely based on visual data coming from a high frame monocular camera.  Inspired by aerial-terrestrial SLAM, we choose to rely on bundle adjustment to optimize the estimated trajectories, thus avoiding the integration of linearization errors of filtered approaches.  Furthermore, our method do not use any environment specific assumption and can hence run in any kind of environment (planar or not).
We show that our method outperforms state-of-the-art visual SLAM algorithms on underwater datasets. 

\section{Features Tracking Methods Evaluation}
\label{sec:track_eval}

As discussed in the introduction, underwater images are mainly degraded by turbidity.  Moreover, underwater scenes do not provide many discriminant features and often show repetitive patterns like coral branches, holes made by animals in the sand or simply algae or sand ripples in shallow waters.  In order to develop a VO system robust to these visual degradation, we have evaluated the performance of different combinations of detectors and descriptors along with the optical flow based \textit{Kanade-Lucas-Tomasi} (KLT) method \cite{KLT_Bouguet} on two sets of underwater images. \\

\begin{figure*}[!t]
	\centering{
    \includegraphics[width=1.\columnwidth]{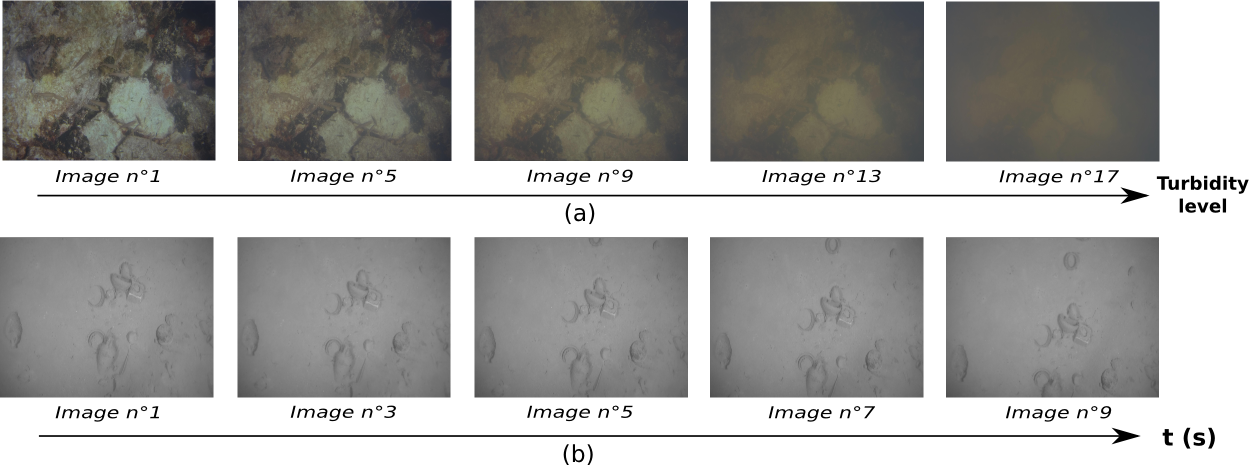}
    }
	\caption{\small Images used to evaluate the tracking of features. (a) Images from the TURBID dataset \cite{Turbid_Dataset_2015}. (b) Images acquired on a deep antic shipwreck (depth: 500 meters, Corsica, France) - Credit: DRASSM (French Department of Underwater Archaeological Research).}
    \label{fig:turbid_and_blade_data}
\end{figure*}

\noindent \textbf{Underwater sets of images.}  Two different sets of images are used here (Fig.                \ref{fig:turbid_and_blade_data}).  The first one is the TURBID dataset \cite{Turbid_Dataset_2015}, which consists of series of static pictures of a printed seabed taken in a pool.  Turbidity was simulated on these images by adding a controlled quantity of milk between two shots.  The second one consists of a sequence of images extracted from a video sequence recorded by a moving camera close to the seabed.  This sequence exhibits the typical characteristics of underwater images: low texture and repetitive patterns.  As this set is a moving one, we will refer to it as the VO set.  On both sets, all the images used are resized to 640x480 and gray-scaled to fit the input format of classical VO methods. \\

\noindent \textbf{Features tracking methods.}  We evaluate the performance of different features tracking methods using handcrafted features only as we seek real-time capacity on CPU, which is not really possible yet with deep learning based methods.  The following combination of features detector and descriptor are compared: ORB \cite{Orb}, BRISK \cite{Brisk}, FAST \cite{Fast} + BRIEF \cite{Brief}, FAST + FREAK \cite{Freak}, SIFT \cite{Sift} and SURF \cite{Surf}.  SIFT and SURF are used as a baseline here as their computational complexity make them unsuitable for real-time applications on CPU.  In addition to these descriptor based methods, we also evaluate the performance of the KLT, an optical flow based method.  The KLT works by extracting Harris corners using the algorithm of \citet{GFTT} and tracking these corners through optical flow with the Lucas-Kanade implementation \cite{lucas-kanade-20years}.  As we will show that the KLT performs best, we also evaluate all the previous descriptors in conjunction with this Harris corner detector for a fair comparison.  \\

\noindent \textbf{Evaluation protocol.}  The TURBID dataset is used to evaluate robustness to turbidity.  The employed evaluation protocol on this set is the following:
\begin{itemize}
\item we divide each image into 500 cells and try to extract one feature per cell
\item we track the features extracted in one image into the following one (\textit{i.e.} the image shot right after an adding of milk)
\item before each tracking, the second image is virtually translated of 10 pixels to avoid initializing the KLT at the right spot
\end{itemize}

\begin{figure}[!t]
	\centering{
    \includegraphics[width=0.75\columnwidth]{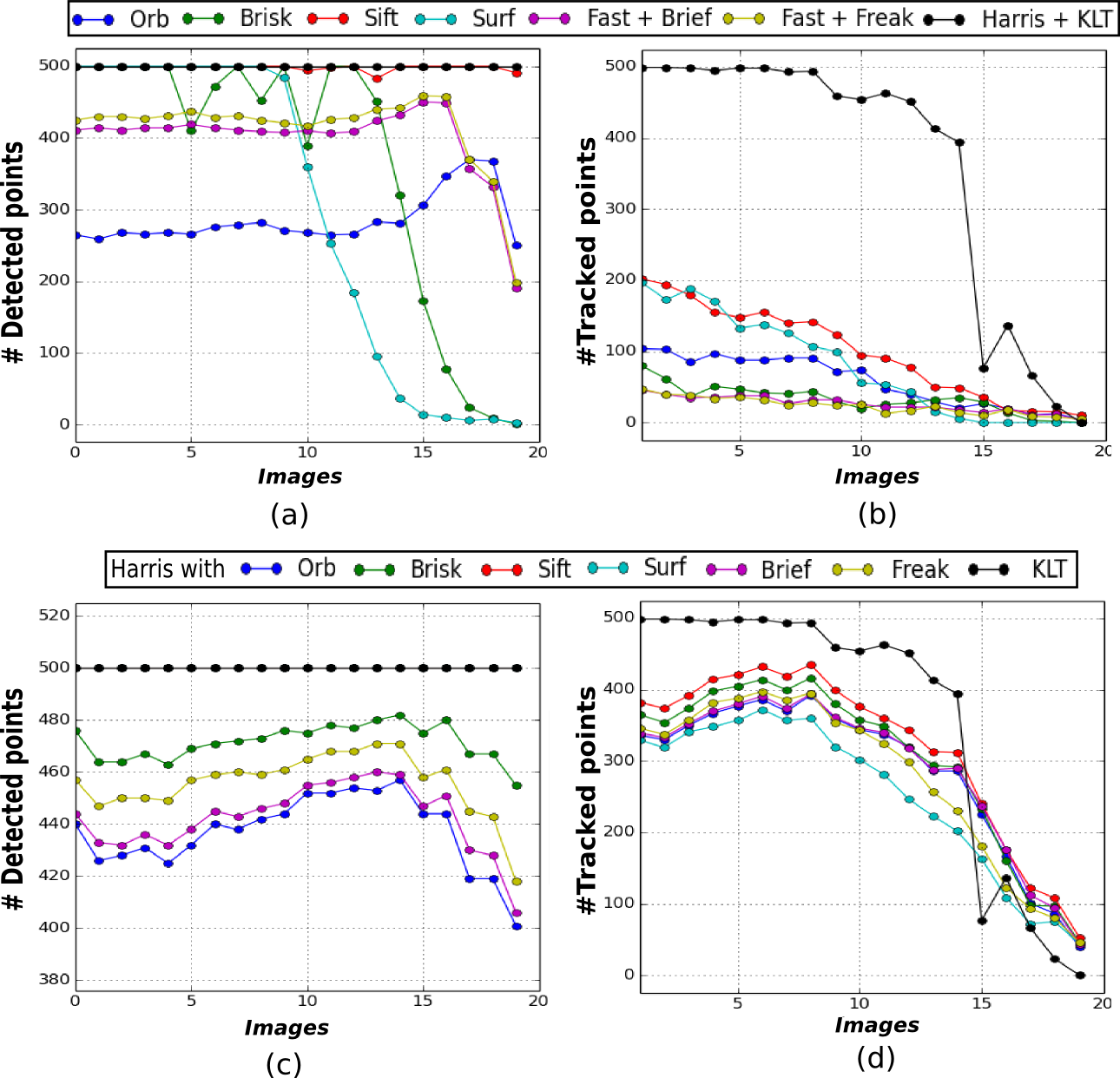}
    }
    \setlength{\belowcaptionskip}{-5pt}
	\caption{\small Features tracking methods evaluation on the TURBID dataset \cite{Turbid_Dataset_2015} (presented in Fig.\ref{fig:turbid_and_blade_data} (a)). Graphs (a) and (b) illustrates number of features respectively detected and tracked with different detectors while (c) and (d) illustrates number of features respectively detected with the Harris corner detector and tracked as before (the SURF and SIFT curves coinciding with the Harris-KLT one in (c)).}
    \label{fig:result_eval_turbid}
\end{figure}

\noindent Note that the KLT method uses a local window, limiting the search space around the previous feature.  As the tracking is here performed between images slightly translated one from the other, the search space for matching descriptors is limited to a 40x40 pixels window around the previous features.  Therefore none of the tested methods is advantaged in front of another.  Note that, as translations are the predominant motion in frame-to-frame tracking in the context of VO, we did not apply rotation or scale change to the images. \\

\noindent The VO set is used to evaluate each method on a real VO scenario, that is we evaluate the efficiency of each method in tracking features over a sequence of images.  For the methods relying on descriptors, we proceed as follows:
\begin{itemize}
\item we divide each image into 500 cells and try to extract one feature per cell
\item we try to match the features extracted in the first image in all the following ones
\end{itemize}

\begin{figure}[t]
	\centering{
    \includegraphics[width=0.75\columnwidth]{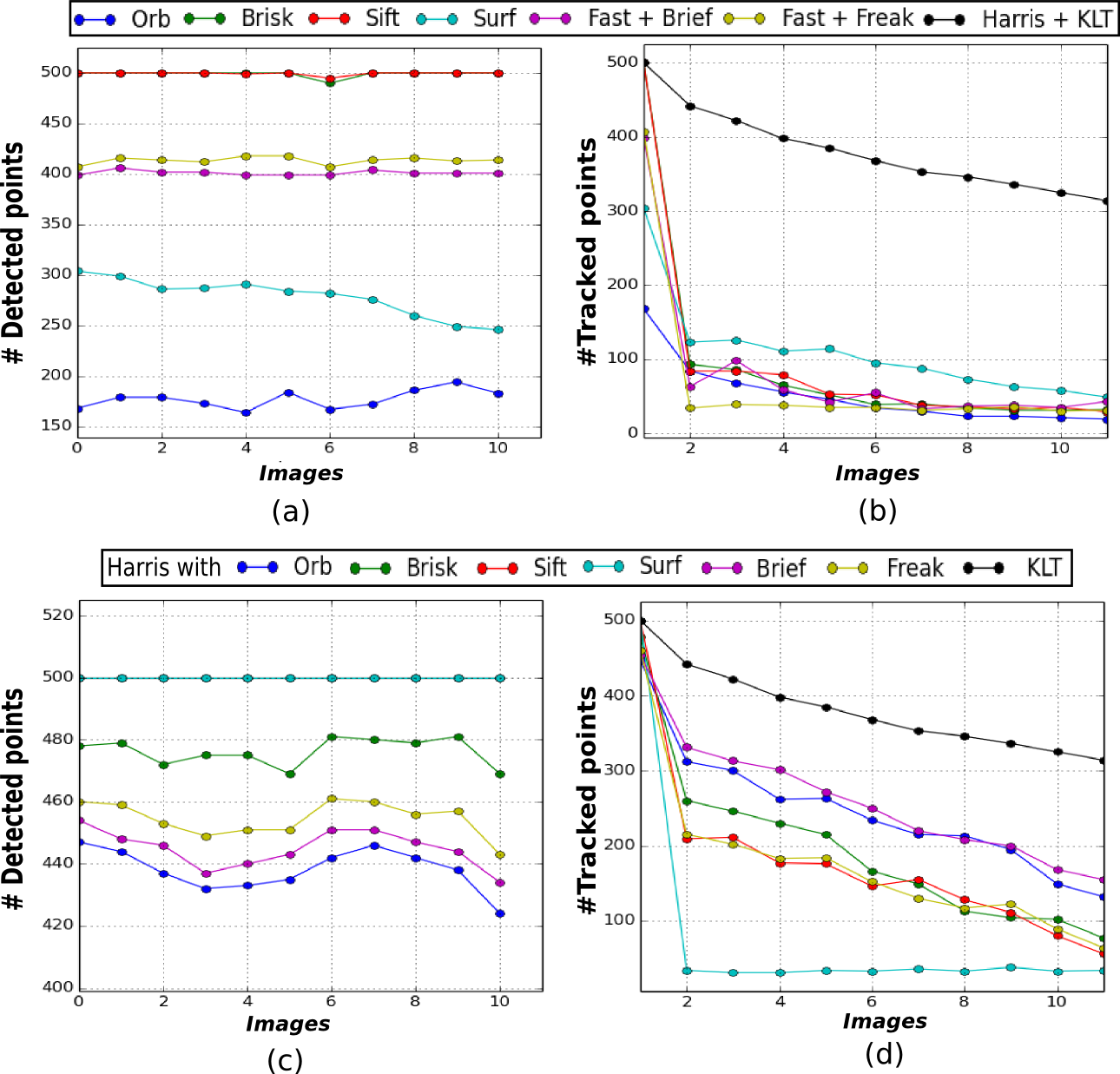}
    }
    \setlength{\belowcaptionskip}{-5pt}
	\caption{\small Evaluation of features tracking methods on a real underwater sequence (presented in Fig.\ref{fig:turbid_and_blade_data} (b)). Graphs (a) and (b) illustrate the number of features respectively detected and tracked with different detectors, while (c) and (d) illustrate the number of features respectively detected with the Harris corner detector and tracked as before (the SIFT curve coinciding with the SURF one in (c)).}
    \label{fig:result_eval_uw}
\end{figure}

\noindent For the KLT method, as it is a local method, we proceed slightly differently:

\begin{itemize}
\item we divide the first image into 500 cells and try to extract one feature per cell
\item we try to track these features sequentially (image-to-image) by computing optical flow in a forward-backward fashion and remove features whose deviation is more than 2 pixels
\end{itemize}

For both methods we removed the outliers by checking the epipolar consistency of matched features in a RANSAC \cite{Ransac} scheme.  On both sets, evaluation metric is the number of correctly tracked features.  Note that, depending on the feature detector, not all methods are be able to detect 500 features in the images. \\

\textbf{Results.}  Fig. \ref{fig:result_eval_turbid} illustrates the results obtained on the TURBID set of images.  Fig. \ref{fig:result_eval_turbid} displays (a) the number of features detected in each image for every method and (b) the number of tracked features between consecutive pictures.  The resulting graphs clearly show that the KLT method is able to track the highest number of features.  Indeed, more than 80\% of the detected features are successfully tracked in the first fifteen images, whereas for the other methods this number is way below 50\%.  However, we can see that the Harris detector is the only one able to extract almost 500 features in each image (Fig. \ref{fig:result_eval_turbid} (a)).  We have therefore run another evaluation using only this detector.  Note that the requirements of some descriptors discard non suited detections, which is the reason of the difference in the number of detected features in Fig. \ref{fig:result_eval_turbid} (c).  The results in Fig. \ref{fig:result_eval_turbid} (d) show that the Harris detector increases the performance of all the descriptors evaluated but none of them matches the performance of the KLT method.

Fig. \ref{fig:result_eval_uw} illustrates the results obtained on the VO set of images.  Fig. \ref{fig:result_eval_uw} (a) and (b) show that the KLT also tracks the highest number of features across this sequence.  Around 60\% of the features detected in the first image are still successfully tracked in the last image with the KLT.  For the other methods, the ratio of features correctly tracked between the first and last image barely reaches 20\%.  Once again, using the Harris detector improves the results for most of the descriptors, increasing the tracking ratio up to about 35\%, but the KLT remains the most efficient tracking method (Fig. \ref{fig:result_eval_uw} (c,d)).

In front of these results, it appears that the KLT method is more robust to the low quality of underwater images.  The reason is that the low texture of the images leads to the extraction of ambiguous descriptors which cannot be matched robustly whereas the KLT, by looking for the minimization of a photometric cost, is less subject to these ambiguities.

Therefore, we choose to build our VO algorithm on the tracking of Harris corners detected with the Shi-Tomasi detector and tracked through optical flow with the method of Lucas-Kanade.  Furthermore, an advantage of using the KLT over using descriptors resides in its low computation cost as it does not require the extraction of new features in each image.

\section{The Visual Odometry Framework}
\label{sec:vo_framework}

\begin{figure}[!t]
	\centering{
    \includegraphics[width=0.50\columnwidth]{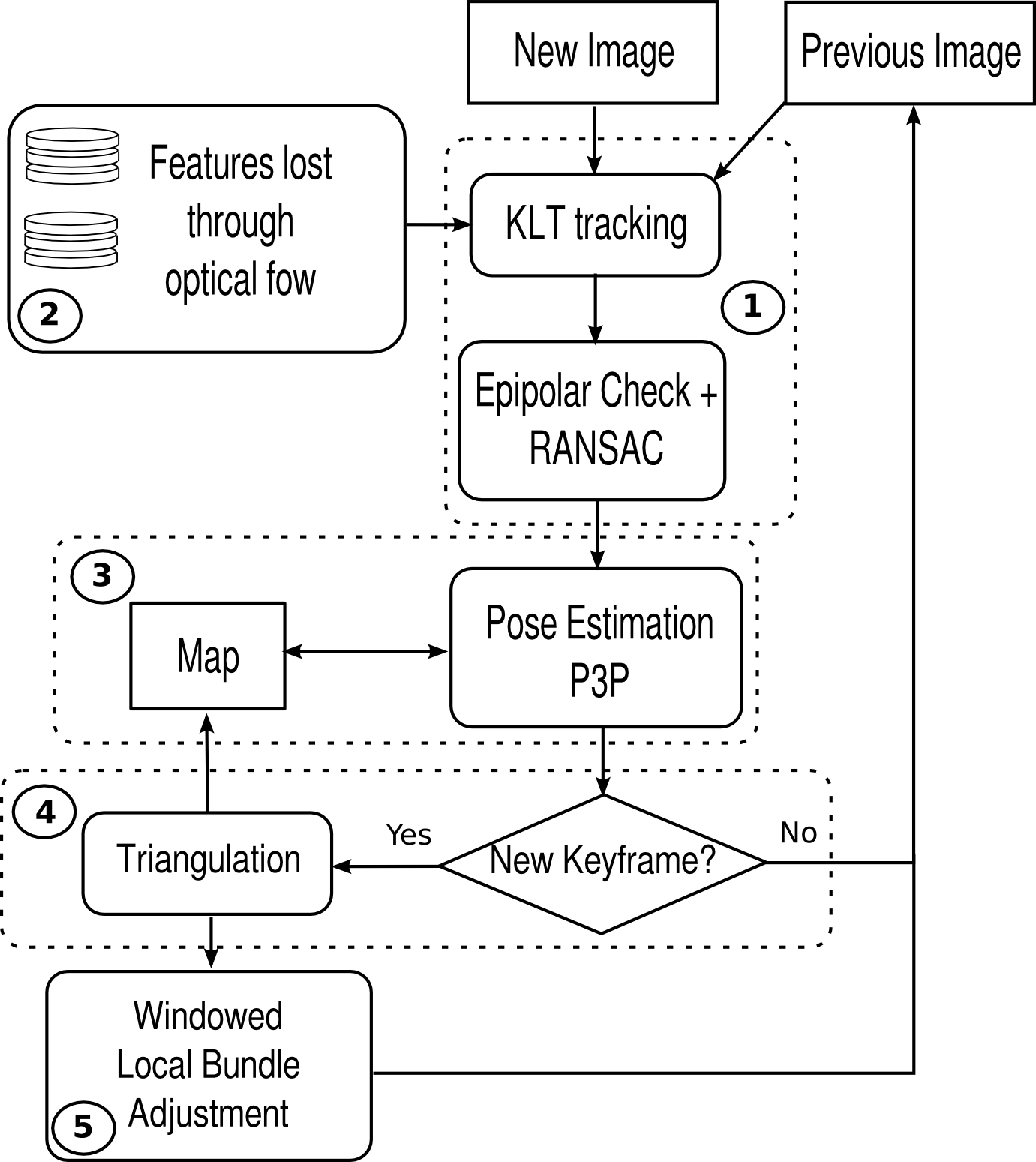}
    }
    \setlength{\belowcaptionskip}{-5pt}
	\caption{\small Pipeline of the proposed visual odometry algorithm.}
    \label{fig:odometry_pipeline}
\end{figure}

The pipeline of UW-VO is summarized in Fig. \ref{fig:odometry_pipeline}.  The system is based on the tracking of 2D features over successive frames in order to estimate their 3D positions in the world referential.  The 2D observations of these 3D landmarks are then used to estimate the motion of the camera.  Frames used for the triangulation of the 3D map points are considered as keyframes and the most recent ones are stored in order to optimize the estimated trajectory along with the structure of the 3D map through bundle adjustment.  The method follows the approach of \cite{PTAM,Strasdat_dwo,ORB-SLAM}.  However, in opposition to these methods, we do not build the tracking on the matching of descriptors.  Instead we use the KLT method, more adapted to the underwater environment as demonstrated in section \ref{sec:track_eval}.  The drawback of the KLT in opposition to descriptors is that it is only meant for tracking features between successive images.  This is a problem when dealing with a dynamic environment as good features might be lost because of short occlusions.  To make the KLT robust to such events, a retracking mechanism is added to the tracking part of the VO framework.  This mechanism will be described in section \ref{sec:retracking}.  Note that, as UW-VO is a fully monocular system, scale is not observable and the camera's pose is hence estimated up to an unknown scale factor.  The recovery of this scale factor could be done by integrating an additional sensor such as an altimeter or an IMU.  This is not considered in this paper but will be the subject of future work. \\

\noindent More formally, the problem of VO can be defined as follow.  At each frame $i$, we want to estimate the state of the system through the pose $\chi_{j}$ of the camera, defined as:
\begin{gather}
\chi_{j} = \begin{bmatrix} p_{j} & q_{j} \end{bmatrix}^{T} \;\;\;\;\;\;\;\;, \;\; with \;\;\; p_{j} \in \mathbb{R}^{3}, \; q_{j} \in SO(3)
\end{gather}
\noindent where $p_j$ is the position of the camera in the 3D world and $q_j$ is its orientation.
\noindent Furthermore, for each newly added keyframe $k$, we want to estimate new landmarks $\lambda_{i}$ $(\lambda_{i} \in \mathbb{R}^{3})$ and then optimize a subset of keyframes pose with the respective observed landmarks.  This set is denoted $\zeta$:
\begin{gather}
\zeta = \left \{ \left. \chi_{k},\chi_{k-1},...,\chi_{k-n},\lambda_{i},...,\lambda_{i-m} \right \} \right.
\end{gather}

\noindent In the following, we assume that the monocular camera is calibrated and that distortion effects are compensated.  The geometrical camera model considered in this work is the pinhole model and its mathematical expression of world points projection is:
\begin{gather} \label{eq:1}
\textbf{proj} (T_{j},X_{i}) = \begin{bmatrix} u \\ v \\ 1 \end{bmatrix} = K . T_{j} . X_{i} \\
\textbf{proj} (T_{j},X_{i}) = \begin{bmatrix} f_x & 0 & c_x \\ 0 & f_y & c_y \\ 0 & 0 & 1 \end{bmatrix} . \begin{bmatrix} R_{j} &  t_{j} \\ 0_{1\times3} & 1 \end{bmatrix} . \begin{bmatrix} x\\ y\\ z\\ 1 \end{bmatrix}
\end{gather}

\noindent with $u$ and $v$ the pixel coordinates, $K$ the intrinsic calibration parameters, $T_{j}$ the projective matrix computed from the state $\chi_{j}$ \(-\) \(T_{j} \in SE(3)\),  \(R_{j} \in SO(3)\), \(t_{j} \in \mathbb{R}^{3}\) \(-\) and \(X_{i} \in \mathbb{P}^{3}\) the homogeneous representation of the landmark $\lambda_{i}$. The coefficients $f_x$,$f_y$ and $c_x$,$c_y$ represent respectively the focal lengths along the $x$ and $y$ axes and the position of the optical center, expressed in pixels. 

\subsection{Frame-to-Frame Features Tracking}
\label{sec:f2f_track}

Features are extracted on every new keyframe using the Shi-Tomasi method to compute Harris corners.  The motion of the 2D features is then estimated using the KLT.  After each optical flow estimation, we thoroughly remove outliers from the tracked features:  first, we compute the disparity between the forward and backward optical flow estimations and remove features whose disparity is higher than a certain threshold.  Then, from the intrinsic calibration parameters, we compute the essential matrix using the 5-points method of \citet{Nister_5pts} between the previous keyframe and the current frame.  This essential matrix is computed within a RANSAC process in order to remove the features not consistent with the estimated epipolar geometry.  

Once enough motion is detected, the tracked 2D features are triangulated from their observations in the current frame and in the previous keyframe.  The current frame used here is converted into a keyframe and new 2D corners are extracted in order to reach a specified maximum number of features.  All these 2D features are then tracked in the same way as described above.

\subsection{Features Retracking}
\label{sec:retracking}

The main drawback of optical flow based tracking is that lost features are usually permanently lost.  In opposition, the use of descriptors allows the matching of features with strong view-point change.  In the underwater context, the powerful lights embedded by ROVs often attract bench of fishes in the camera field of view.  The occlusions due to fishes can lead to strong photometric shifts and consequently to a quick loss of features.  However, fishes are moving very fast and their position changes very quickly between successive frames.  We take advantage of this fact to increase the robustness of our tracking method over short occlusions.  The employed strategy is too keep a small window of the most recent frames (five frames is enough in practice) with the list of features lost through optical flow in it.  At each tracking iteration, we try to retrack the lost features contained in the retracking window.  Finally, retracked features are added to the set of currently tracked features.

This features tracking implementation is used to track both pure 2D features, for future triangulation, and 2D observations of already mapped points, for pose estimations.

\subsection{Pose Estimation}
\label{sec:pose_estim}

The estimation of the 6 degrees of freedom of the pose of every frame uses their respective 2D-3D correspondences.  The pose is computed with the Perpective-from-3-Points (P3P) formula, using the method of \citet{P3P_Kneip}.  This operation is done within a RANSAC loop to remove inaccurate correspondences.  The pose is computed from the combination of points giving the most likely estimation for the set of features.  The pose is then further refined by minimization of the global reprojection error using the set $I$ of inliers: 
\begin{gather} \label{eq:3}
\underset{T_j}{\mathrm{arg\min}} \sum_{i \in I} (x_i - proj(T_j,X_i))^{2}
\end{gather}

\noindent with $x_i$ the 2D observation of the world point $X_i$ and $proj(T_j,X_i)$ the reprojection of $X_i$ in the frame $j$ with its related projection matrix $T_j$.

This minimization is done through a nonlinear least-squares optimization using the Levenberg-Marquardt algorithm.  The computed poses are then used to estimate the 3D positions of the tracked features.

\subsection{Keyframe Selection and Mapping}
\label{sec:kf_selec}

The mapping process is triggered by the need of a new keyframe.  Several criteria have been set as requirements for the creation of a keyframe.  The first criterion is the parallax.  If an important parallax from the last keyframe has been measured (30 pixels in practice), a new keyframe is inserted as it will allow the computation of accurate 3D points.  The parallax is estimated by computing the median disparity of every tracked pure 2D features from the previous keyframe.  To ensure that we do not try to estimate 3D points from false parallax due to rotational motions, we unrotate the currently tracked features before computing the median disparity.  The second criterion is based on the number of 2D-3D correspondences.  We verify that we are tracking enough 2D observations of map points and trigger the creation of a keyframe if this number drops below a threshold defined as less than 50\% of the number of observations in the last keyframe.

For further optimization, a window of the most recent keyframes along with their 2D-3D correspondences is stored.  This optimization operation known as bundle adjustment is performed in parallel after the creation of every keyframe and is described next.  Finally, new Harris corners are detected and the tracking loop is run again.

\subsection{Windowed Local Bundle Adjustment}
\label{sec:BA}

As stated above, a window of the most recent $N$ keyframes is stored and optimized with bundle adjustment at the creation of new keyframes.  To ensure a reasonable computational cost, only a limited number of the most recent keyframes are optimized along with their tracked map points.  The remaining keyframes are fixed in order to constrain this nonlinear least-squares optimization problem.  The size of the window is set adaptively by including every keyframe sharing a map point observation with one of the optimized keyframes.  This adaptive configuration sets high constraints on the problem and helps in reducing the unavoidable scale drift inherent to monocular odometry systems.  The Levenberg-Marquardt algorithm is used to perform this optimization.  The problem is solved by minimizing the map points reprojection errors.  As least-squares estimators do not make any difference between high and low error terms, the result would be highly influenced by the presence of outliers with high residuals.  To prevent this, we use the robust M-Estimator Huber cost function \cite{Hartley2004} in order to reduce the impact of the highest error terms on the found solution.

\noindent We define the reprojection errors $e_{ij}$ for every map point $i$ observed in a keyframe $j$ as:
\begin{gather} \label{eq:4}
e_{ij} = x_{ij} - proj(T_j,X_i)
\end{gather}
\noindent We then define the set of parameters $\zeta^{*}$ to optimize as:
\begin{gather}
\zeta^{*} = \left \{ \left. \chi_{j-2},\chi_{j-1},\chi_{j},\lambda_{i},...,\lambda_{i+M} \right \} \right.
\end{gather}
\noindent with $M$ the number of landmarks observed by the three most recent keyframes. 
\noindent And we minimize \eqref{eq:4} over the optimization window of $N$ keyframes:
\begin{gather} \label{eq:5}
\underset{\zeta^{*}}{\mathrm{arg\min}} \sum_{j \in N} \sum_{l \in L_{j}} \rho (e_{ij}^{T}\Sigma_{ij}^{-1}e_{ij})
\end{gather}

\noindent with $L_{j}$ the set of landmarks observed by the keyframe $j$, $\rho$ the Huber robust cost function and $\Sigma_{ij}$ the covariance matrix associated with the measures $x_{ij}$.

After convergence of the Levenberg-Marquardt algorithm, we remove the map points with a resulting reprojection error higher than a threshold.  This optimization step ensures that after the insertion of every keyframe both the trajectory and the 3D structure of the map are statistically optimal.

\subsection{Initialization}
\label{sec:ini}

Monocular systems are subject to a "Chicken and Egg" problem at the beginning.  Indeed, the motion of the camera is estimated through the observations of known 3D world points, but the depth of the imaged world points is not observable from a single image.  The depth of these world points can be estimated using two images with a sufficient baseline.  However, this baseline needs to be known to compute the depth and vice-versa.  This is why monocular VO requires an initialization step to bootstrap the algorithm in opposition with stereo systems.  In UW-VO, initialization is done here by computing the relative pose between two frames through the estimation of an essential matrix with the 5-points methods of \cite{Nister_5pts}.  The norm of the estimated translation vector is then arbitrarily fixed to one, as scale is not observable with monocular setups.  We assessed that this simple method is able to initialize accurately the VO framework in any configuration (planar or not), making it non-environment dependent. 

\section{Experimental Results}
\label{sec:results}

\textbf{Implementation :} The proposed system has been developed in C++ and uses the ROS middleware \cite{ROS}.  The tracking of features is done with the OpenCV implementation of the Kanade-Lucas algorithm \cite{KLT_Bouguet}.  Epipolar geometry and P3P pose estimations are computed using the OpenGV library \cite{OpenGV}.  Bundle Adjustment is performed using the graph optimization framework g2o \cite{g2o} and runs in a parallel thread.  The average run time is of 25ms per frame with the tracking limited to 250 features per frame and bundle adjustment is performed on the five most recent keyframes.  The run time goes up to 35ms when a new keyframe is required because of the features detection and triangulation overload.  Thus our system can run in real-time for video sequences with a frame rate up to 30 Hz.  The experiments have been carried with an Intel Core i5-5200 CPU - 2.20GHz - 8 Gb RAM.

\vspace{2.5mm}

To the best of our knowledge there is no underwater method able to estimate localization from monocular images available open-source.  Furthermore, no publicly  available datasets were released with these methods, so we cannot compare with them.  Hence, UW-VO has been evaluated along with ORB-SLAM\footnote{https://github.com/raulmur/ORB\_SLAM2}, LSD-SLAM\footnote{https://github.com/tum-vision/lsd\_slam} and SVO\footnote{http://rpg.ifi.uzh.ch/svo2.html} on different datasets which are all available online, allowing future methods to compare to our results.

All algorithms are evaluated on real underwater datasets.  UW-VO and ORB-SLAM are also evaluated on a simulated dataset, whose frame rate (10 Hz) is too low for SVO and LSD-SLAM to work.  Indeed, SVO and LSD-SLAM are direct methods which require very high overlap between two successive images in order to work.  Note that ORB-SLAM and SVO have been fine-tuned in order to work properly.  For ORB-SLAM, the features detection threshold was set at the lowest possible value and the number of points was set to 2000.  For SVO, the features detection threshold was also set at the lowest possible value and the number of tracked features required for initialization was lowered to 50.  For each method, every results presented are the averaged results over five runs.

\subsection{Results on a Simulated Underwater Dataset}
\label{sec:uwsim}

\begin{figure}[!h]

    \centering
    \begin{minipage}{0.45\textwidth}
        \centering
        \includegraphics[width=0.85\textwidth]{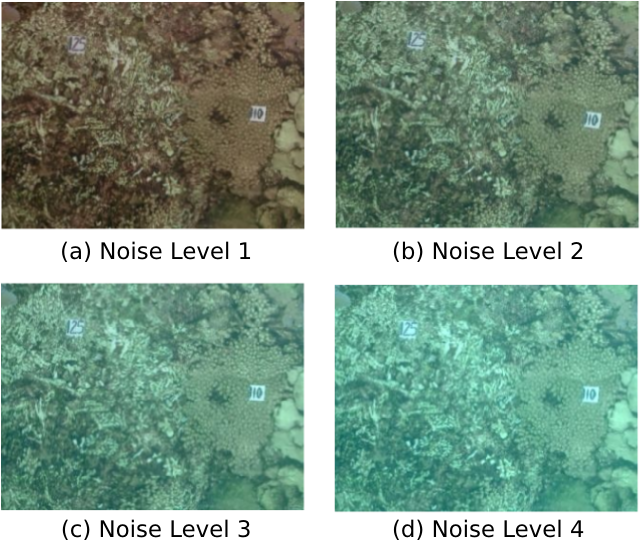} 
        \caption{The four different turbidity levels of the simulated dataset.}
        \label{fig:uwsim_dataset}
    \end{minipage}\hfill
    \begin{minipage}{0.45\textwidth}
        \centering
        \includegraphics[width=1.15\textwidth]{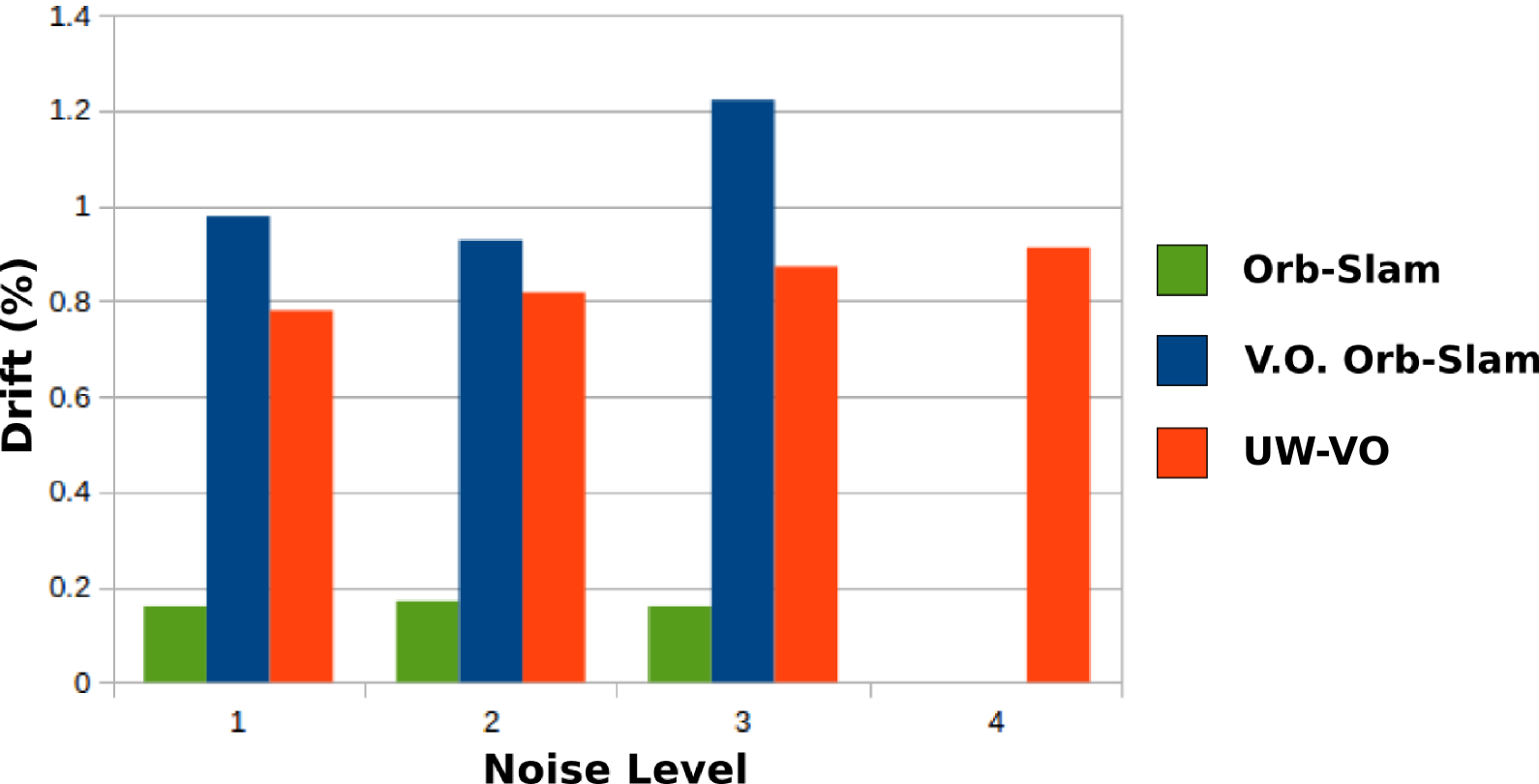} 
        \caption{Drift of ORB-SLAM (green), V.O. ORB-SLAM (blue) and UW-VO (red) on the simulated underwater dataset.}
        \label{fig:uwsim_results}
    \end{minipage}

\end{figure}

A simulated dataset created from real underwater pictures has been made available to the community by \citet{UWsimData}.  Four monocular videos of a triangle-shaped trajectory are provided with four different levels of noise in order to synthetically degrade the images with turbidity-like noise (Fig \ref{fig:uwsim_dataset}).  The images resolution of these videos is 320x240 pixels.  In each sequence, the triangle-shaped trajectory is performed twice and it starts and ends at the same place.  These four sequences have been used to evaluate the robustness against turbidity of UW-VO with respect to ORB-SLAM.  For fair comparison, ORB-SLAM has been run with and without its loop-closing feature.  We will refer this version of ORB-SLAM as V.O. ORB-SLAM in the following. 

Table \ref{table:uwsim_results} presents the final drift at the end of the trajectory for each method.  On the first three sequences, ORB-SLAM is able to close the loops and therefore has the lowest drift values, as the detection of the loop closures allows to reduce the drift accumulated in-between.  On the same sequences, V.O. ORB-SLAM has the highest level of drifts.  Note that ORB-SLAM and its V.O. alternative fail half the time on the third level of noise sequence and have been run many times before getting five good trajectories.  It is worth noting that the localization drift increases significantly for V.O. ORB-SLAM when the turbidity level gets higher.  This is mostly due to the increased inaccuracy in its tracking of ORB features.  On the last sequence, the turbidity level is such that ORB descriptors get too ambiguous and leads to failure in ORB-SLAM tracking.  These results highlight the deficiency of ORB-SLAM tracking method on turbid images.  
In comparison, UW-VO is able to run on all the sequences, including the ones with the highest levels of noise (Fig. \ref{fig:uwsim_results}).  The computed trajectories are more accurate than V.O. ORB-SLAM and we can note that it is barely affected by the noise level.  These results confirm the efficiency of UW-VO as a robust odometry system in turbid environments.

\begin{table}[!h]
\centering
\begin{tabular}{@{}ccccc@{}}
\toprule  
& & \multicolumn{3}{c}{\textbf{Drift (in \%)}}  

\\ \cmidrule(l){3-5}

\multicolumn{1}{l}{Seq. Noise Level} & Turbidity & \textit{ORB-SLAM} & V.O. ORB-SLAM & UW-VO         
 
 \\ \midrule \midrule

\textit{1}                                               & None                                                        & \textit{0.18}     & 0.97          & \textbf{0.78} \\ \hline

{\textit{2}}                                                 & Low                                                         & \textit{0.18}     & 0.93          & \textbf{0.81} \\ \hline

{\textit{3}}                                                 & Medium                                                      & \textit{0.17*}     & 1.21*          & \textbf{0.85} \\ \hline

{\textit{4}}                                                 & High                                                        & \textit{X}        & X             & \textbf{0.89} \\ \hline

\end{tabular}
\caption{Translation drift (in \%) on the simulated underwater video sequence with different level of noise simulating turbidity effects.  Results are given averaging over five runs for each algorithm.  V.O. ORB-SLAM designates ORB-SLAM without the loop closing feature enabled, i.e. performing only Visual Odometry.  ORB-SLAM results are given for information. The (*) denotes very frequent failure of the algorithm.}
\label{table:uwsim_results}
\end{table}

\begin{figure}[!h]
	\centering{
    \includegraphics[width=.75\textwidth]{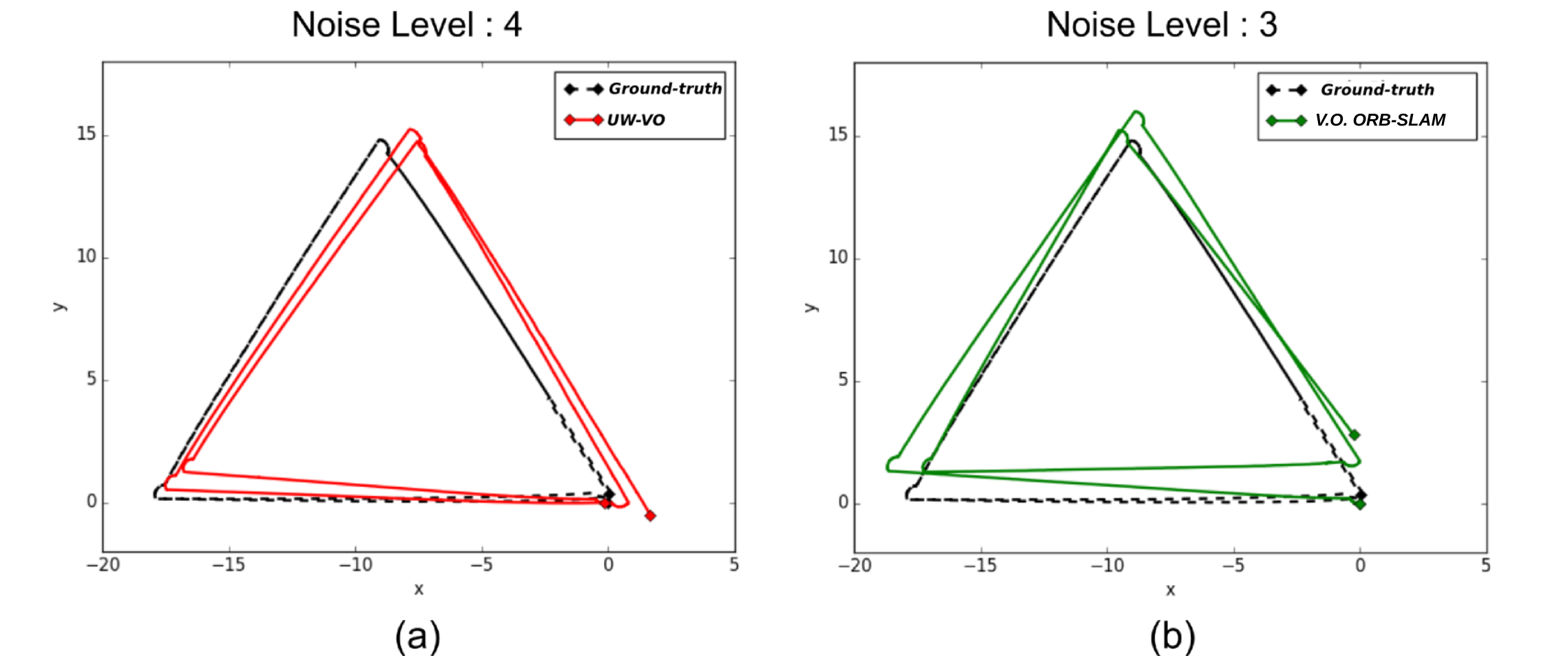}
    }
	\caption{\small Trajectories estimated with (a) our method on the sequence with the highest level of noise and with (b) V.O. ORB-SLAM on the sequence with the noise level of 3.}
    \label{fig:uwsim_traj}
\end{figure}

\subsection{Results on a Real Underwater Video Sequence}
\label{sec:real_uw_data}

We now present experiments conducted on five real underwater video sequences.  These sequences were gathered 500 meters deep in the Mediterranean Sea (Corsica), in 2016, during an archaeological mission conducted by the French Department of Underwater Archaeological Research (DRASSM).  The videos were recorded from a camera embedded on an ROV and gray-scale 640x480 images were captured at 16 Hz.  The calibration of the camera has been done with the Kalibr \cite{Kalibr} library.  Calibration was done in situ in order to estimate the intrinsic parameters and the distortion coefficients of the whole optical system.  If the calibration is performed in the air, the water and camera's housing effects on the produced images would not be estimated without simulating their effects \cite{Luczynski17} and this would lead to a bad estimate of the camera's parameters.  The camera recording the videos was placed inside an underwater housing equipped with a spherical dome and we obtained good results using the pinhole-radtan model (assessed by a reprojection error $<$ 0.2 px).

\noindent These five sequences can be classified as follow:

\begin{itemize}
\item Sequence 1: low level of turbidity and almost no fishes.
\item Sequence 2: medium level of turbidity and some fishes.
\item Sequence 3: high level of turbidity and many fishes.
\item Sequence 4: low level of turbidity and many fishes.
\item Sequence 5: medium level of turbidity and many fishes.
\end{itemize}

For each of these sequences, a ground truth was computed using the state-of-the-art Structure-from-Motion software Colmap \cite{Colmap}.  Colmap computes trajectories offline by exhaustively trying to match all the images of a given sequence, thus finding many loops and creating very reliable trajectories.  We could assess the accuracy of the reconstructed trajectories both visually and by checking the correctness of the matched images.  

\begin{figure}[!t]
	\centering{
    \includegraphics[width=1.\textwidth]{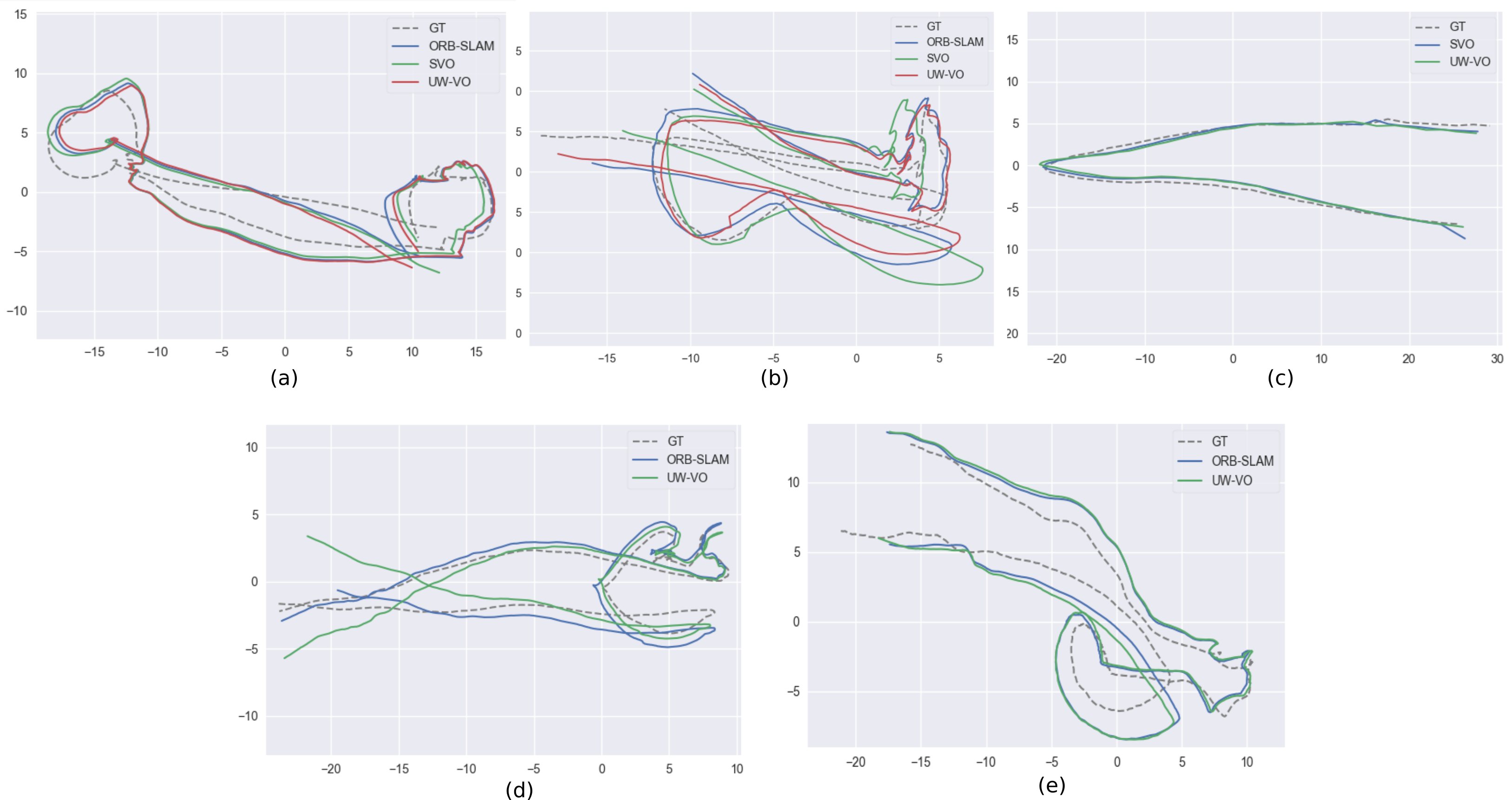}
    }
	\caption{\small Trajectories of ORB-SLAM, SVO and UW-VO over the five underwater sequences. (a) Sequence 1, (b) Sequence 2, (c) Sequence 3, (d) Sequence 4, (e) Sequence 5. Ground-truths (GT) are extracted from Colmap trajectories.}
    \label{fig:blade_trajs}
\end{figure}

Here, we compare ORB-SLAM, LSD-SLAM and SVO to UW-VO.  We evaluate the results of each algorithm against the trajectories computed offline by Colmap by first aligning the estimated trajectories with a similarity transformation using the method of \cite{Umeyama} and then computing the absolute trajectory error \cite{Sturm_ATE} (Fig. \ref{fig:blade_trajs}).  The results are displayed in table \ref{table:blade_results}.  To observe the effect of the retracking mechanism (described in section \ref{sec:retracking}), we have run the UW-VO algorithm with and without enabling this feature, respectively referring to it as UW-VO and UW-VO$^{*}$ (\textit{Videos of the results for each method on the five sequences are available online\footnote{https://www.youtube.com/playlist?list=PL7F6c8YEyil-RuC7YptNMAM88gfBfn0u4}}).

\begin{table}[!t]
\centering
\begin{tabular}{@{}ccccccccc@{}}
\toprule

& & & & \multicolumn{5}{c}{\textbf{Absolute Trajectory Error RMSE( in \%)}} 
\\
\cmidrule(l){5-9}

Seq. \# & Duration & \begin{tabular}[c]{@{}c@{}} Turbidity \\ Level \end{tabular} & \begin{tabular}[c]{@{}c@{}} Short \\ Occlusions \end{tabular} & LSD-SLAM & ORB-SLAM & SVO & UW-VO* & UW-VO
\\ \midrule \midrule

1  & \textit{4'}     & \textit{Low}                                              & \textit{Few}                                               & X        & 1.67          & \textbf{1.63} & 1.78  & 1.76          \\ \hline
2  & \textit{2'30''} & \textit{Medium}                                           & \textit{Some}                                              & X        & 1.91          & 2.45         & 1.78  & \textbf{1.73} \\ \hline
3  & \textit{22''}   & \textit{High}                                             & \textit{Many}                                              & X        & X              & 1.57         & 1.10  & \textbf{1.04} \\ \hline
4  & \textit{4'30''} & \textit{Low}                                              & \textit{Many}                                              & X        & \textbf{1.13} & X             & 1.61  & 1.58          \\ \hline
5  & \textit{3'15''} & \textit{Medium}                                           & \textit{Many}                                              & X        & 1.94           & X             & 2.08  & \textbf{1.88} \\ \hline
\end{tabular}
\caption{Absolute translation errors (RMSE in \%) for five underwater sequences with different visual degradation.  Results are given averaging over five runs for each algorithm.  UW-VO* designates our method without the retracking step, while UW-VO designates our method with the retracking step.}
\label{table:blade_results}
\end{table}

As we can see, LSD-SLAM fails on all the sequences.  This is most likely due to its semi-dense approach based on the tracking of edges with strong gradients, which are not frequent on sea-floor images.  SVO is able to compute quite accurate trajectories on the sequences that are not too much affected by dynamism from moving fishes.  The tracking of SVO, which is similar to optical flow, seems to work well even on turbid images, but its direct pose estimation method is not robust to bad tracked photometric patches like the one created by moving fishes (seq. 3,4,5).  ORB-SLAM on the other hand performs well on highly dynamic sequences, but loses in accuracy when turbidity is present (seq. 2,3,5).  Its pose estimation method based on the observations of independent features is hence robust to short occlusions and dynamic objects, but its tracking method fails on images degraded by turbidity.  Furthermore, we can note that despite loop closures in the trajectories (see Fig. \ref{fig:blade_trajs}), ORB-SLAM is not able to detect them.  The failure to detect the loop closures indicates that the Bag of Words approaches \cite{DBOW} might not be suited to the underwater environment, which does not provide many discriminant features.

UW-VO is the only method able to run on all the sequences.  While the estimated trajectory is slightly less accurate on the easiest sequence (seq. 1), UW-VO performs better than ORB-SLAM and SVO on the hardest sequences (seq. 2,3,5, with turbidity and dynamism, which is very common during archaeological operations).  We can see the benefit of the developed retracking mechanism on most of the sequences.  Nonetheless, this optical flow retracking step is not as efficient as the use of descriptors when the number of short occlusions is very large (seq. 4).  Studying the effect of combining optical flow tracking with the use of descriptors could result in an interesting hybrid method for future work.

\section{Conclusion} 
\label{sec:conclusion}

In this paper we have presented UW-VO, a new vision-based underwater localization method.  While most of the existing approaches rely on expensive navigational sensors to estimate the motions of underwater vehicles, we have chosen to investigate the use of a simple monocular camera as a mean of localization.  We propose a new keyframe based monocular visual odometry method robust to the underwater environment.  Different features tracking methods have been evaluated in this context and we have shown that optical flow performs better than the classical methods based on the matching of descriptors.  We further enhanced this optical flow tracking by adding a retracking mechanism, making it robust to short occlusions due to the environment dynamism.  We have shown that the proposed method outperforms the state-of-the-art visual SLAM algorithms ORB-SLAM, LSD-SLAM and SVO in underwater environments.  We publicly released the underwater datasets used in this paper along with the camera calibration parameters and the trajectories computed with Colmap to allow future methods to compare to our results.
The good results obtained on these sequences highlight the effectiveness of the developed method for localizing ROVs navigating in deep underwater archaeological sites.  The computed localization could be used by the pilot as a driving assistance and could further serves as a feedback information for navigation if scale is recovered.  Future work will study the implementation of this localization algorithm on an embedded computing unit in order to fulfill these tasks.
The development of a monocular visual odometer was a first step towards a robust underwater localization method from low-cost sensors.  One perspective is to enhance it by adding a loop-closure mechanism, turning into a visual SLAM method.  We have observed that loop-closing approaches based on classical Bag of Words \cite{DBOW} do not work as expected results in our tests and alternative methods in the lead of \cite{ClusterBased_UW_LC} need to be investigated.  Finally, in the same idea as visual-inertial SLAM algorithms, we will next study the tight fusion of a low-cost IMU and of a pressure sensor with this visual method to improve the localization accuracy and retrieve the scale factor.


\vspace{6pt} 



\authorcontributions{M.F. conducted this research during his doctoral work. M.F. designed and implemented the UW-VO algorithm.  M.F., J.M. and P.T. analyzed the results of UW-VO.  V.C. did the acquisition of the video sequences used.  All the authors participated in the redaction of the article.}

\funding{This research received no external funding.}

\acknowledgments{The authors acknowledge support of the CNRS (Mission pour l'interdisciplinarit\'e - Instrumentation aux limites 2018 - Aqualoc project) and support of R\'egion Occitanie (ARPE Pilotplus project).  The authors are grateful to the DRASSM for its logistical support and for providing the underwater video sequences.}

\conflictsofinterest{The authors declare no conflict of interest.} 

\reftitle{References}


\externalbibliography{yes}
\bibliography{references}



\end{document}